\documentclass[10pt,twocolumn,letterpaper]{article}

\usepackage{iccv}
\usepackage{times}
\usepackage{epsfig}
\usepackage{graphicx}
\usepackage{amsmath}
\usepackage{amssymb}

\usepackage[pagebackref=true,breaklinks=true,letterpaper=true,colorlinks,bookmarks=false]{hyperref}

\usepackage[capitalize]{cleveref}
\crefname{section}{Sec.}{Secs.}
\Crefname{section}{Section}{Sections}
\Crefname{table}{Table}{Tables}
\crefname{table}{Tab.}{Tabs.}

\usepackage{subcaption}
\usepackage{enumitem} 
\usepackage{booktabs}
\usepackage{multirow}
\usepackage{rotating}
\usepackage{adjustbox}

\def\ie{\emph{i.e}\onedot} 

\usepackage{mathtools}
\usepackage{textcomp}
\usepackage{gensymb}
\usepackage[dvipsnames]{xcolor}

\newcommand{\quotes}[1]{``#1''}
\newcommand{\myparagraph}[1]{\vspace{4pt}\noindent\textbf{#1.}}
\newcommand\our{Divide\&Classify}
\newcommand\ours{D\&C}

\iccvfinalcopy 

\def\iccvPaperID{8939} 

\ificcvfinal\fi

\begin{document}

\title{Divide\&Classify: Fine-Grained Classification for\\ City-Wide Visual Place Recognition}

\author{Gabriele Trivigno*$^{1}$
\quad
Gabriele Berton*$^{1}$
\quad
Juan Aragon$^{1}$
\quad
Barbara Caputo$^{1}$
\quad
Carlo Masone$^{1}$\\
$^{1}$ Politecnico di Torino\\
{\tt\small \{gabriele.trivigno, gabriele.berton, barbara.caputo, carlo.masone\}@polito.it}\\
}

\maketitle
\ificcvfinal\thispagestyle{empty}\fi

\begin{abstract}
Visual Place recognition is commonly addressed as an image retrieval problem. However, retrieval methods are impractical to scale to large datasets, densely sampled from city-wide maps, since their dimension impact negatively on the inference time. Using approximate nearest neighbour search for retrieval helps to mitigate this issue, at the cost of a performance drop.
In this paper we investigate whether we can effectively approach this task as a classification problem, thus bypassing the need for a similarity search.
We find that existing classification methods for coarse, planet-wide localization are not suitable for the fine-grained and city-wide setting. This is largely due to how the dataset is split into classes, because these methods are designed to handle a sparse distribution of photos and as such do not consider the visual aliasing problem across neighbouring classes that naturally arises in dense scenarios.
Thus, we propose a partitioning scheme that enables a fast and accurate inference, preserving a simple learning procedure, and a novel inference pipeline based on an ensemble of novel classifiers that uses the prototypes learned via an angular margin loss. 
Our method, Divide\&Classify (D\&C), enjoys the fast inference of classification solutions and an accuracy competitive with retrieval methods on the fine-grained, city-wide setting. 
Moreover, we show that D\&C can be paired with existing retrieval pipelines to speed up computations by over 20 times while increasing their recall, leading to new state-of-the-art results.
Code is available at {\small{\url{https://github.com/ga1i13o/Divide-and-Classify}}}
\end{abstract}


\section{Introduction}
\label{sec:introduction}

Visual Place Recognition (VPR) is the task of recognizing the location where a photo was taken with an accuracy of a few meters
\cite{Torii-2015,Arandjelovic_2018_netvlad, Kim_2017_crn,Torii_2018_tokyo247,Liu_2019_sare, Doan-2019, Ge_2020_sfrs, Berton_2021_svox, Ibrahimi_2021_insideout_vpr, Warburg_2020_msls,seqvlad,Hausler_2021_patch_netvlad,Torii_2021_r_sf,Zaffar-2021,Chen-2017,Chen-2017b,Chen-2018b,Garg-2019,Khaliq-2020,Hausler-2019}. This problem, also known as visual geo-localization  \cite{Berton_CVPR_2022_cosPlace, Berton_2022_benchmark, Kim_2017_crn} or image localization \cite{Liu_2019_sare, Ge_2020_sfrs}, 
is commonly approached as an image retrieval problem: the query to be localized is compared to a database of geo-tagged images, typically via a k-nearest neighbour (kNN) in features space, and the most similar images retrieved from the database are the predictions of the query's location. 
Even though retrieval methods work remarkably well when the VPR task is limited to a small map~\cite{Berton_2022_benchmark, Zaffar-2021, Berton_CVPR_2022_cosPlace}, scaling them to large and densely mapped areas, such as an entire city, is impractical because both the time and memory required to execute the kNN grow with the dimension of the database~\cite{Zaffar-2021, Berton_2022_benchmark, Pion-2020, Babenko_2012_inv_multiindex}.

\begin{figure}
    \begin{center}
    \includegraphics[width=\linewidth]{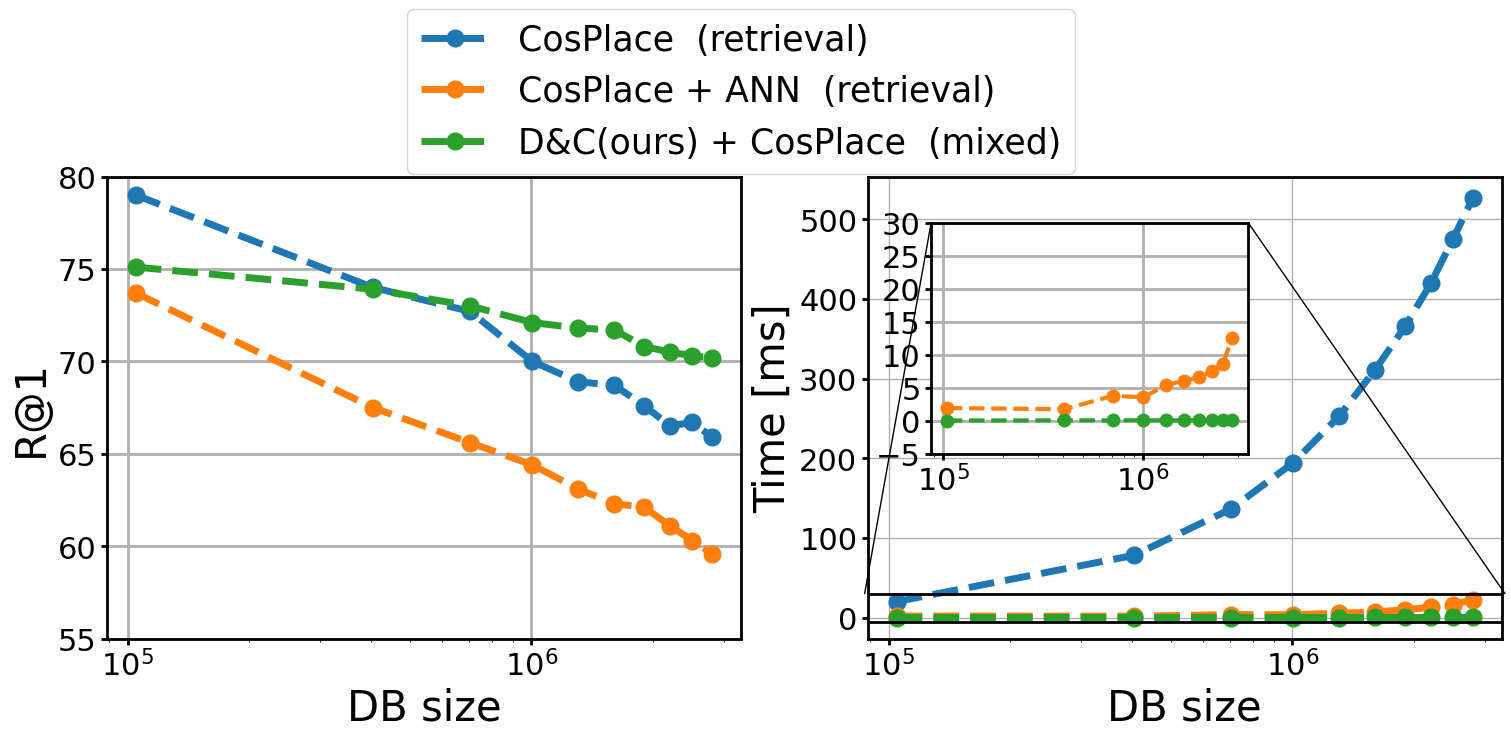}
    \end{center}
    \caption{\textbf{Experiments} demonstrating the scalability problem of retrieval-based VPR methods, using the state-of-the-art CosPlace (with exhaustive kNN and with Approximate Nearest Neighbor - ANN - search through Inverted File Index with Product Quantization, IVFPQ). Combining our Divide\&Classify method with the retrieval approach yields an optimal performance-efficiency trade-off when scaling up to the city-wide setting.
    }
    \label{fig:teaser_new}
\end{figure}


Certainly, the space and time requirements of the kNN can be reduced by resorting to Approximate Nearest Neighbors (ANNs) algorithms \cite{Jegou_2011_productQ, Malkov_2020_hnsw, Babenko_2012_inv_multiindex, Sivic_2003_ivf}, which grant impressive speed ups while simultaneously reducing memory footprint. However, their accuracy may be significantly worse than the one obtained with the exhaustive kNN, which is their upper bound.
This trade-off is easily demonstrated with an experiment. In \cref{fig:teaser_new} we show the behavior obtained using the state-of-the-art retrieval method CosPlace~\cite{Berton_CVPR_2022_cosPlace} on SF-XL \cite{Berton_CVPR_2022_cosPlace}, a dataset for urban VPR that covers the whole city of San Francisco, with an area of nearly 170km$^2$ and over 40M images. Starting from a reduced version of SF-XL and gradually increasing the size of the database up to its full version, we can observe that the inference time quickly explodes with the exhaustive kNN. On the other hand, with an ANN the inference time grows much more slowly (yet still linearly \wrt the size of the database), at the cost of a big drop in accuracy.

A different route to address the scalability problem is to frame VPR as a classification task, so that predictions of the query's location can be obtained without a similarity search over a database.
So far, this formulation has been applied in the setting of planet-wide, coarse geolocalization
\cite{Muller_2018_hierarchical_geolocation, Weyand_2016_PlaNet, Seo_2018_CPlaNet, Kordopatis_2021_EfficientNetGeoloc}. 
All these methods are designed to divide the globe in very coarse classes, spanning up to hundreds of kilometers each and following the sparse and uneven distribution of photos in the planet.
We question whether these classification approaches can be adapted to the fine-grained city-scale VPR setting, where the required localization accuracy is in the order of a few meters.
We find that the global-scale classification methods~\cite{Muller_2018_hierarchical_geolocation, Weyand_2016_PlaNet, Seo_2018_CPlaNet, Kordopatis_2021_EfficientNetGeoloc}, albeit being much faster than their retrieval counterparts, are unsuitable for this setting because they are far too imprecise and because they do not account for the visual overlap among classes that arises when the photos are densely sampled from the map.

Therefore, we propose a novel classification-based VPR technique that is specifically designed for the dense and homogeneous configurations of large urban areas. Our method,  called \textbf{\our} (D\&C), has the speed advantage of classification approaches while being competitive with retrieval methods in terms of accuracy.
Most importantly, we demonstrate that the predictions from D\&C can be used to restrict the search space of retrieval methods, combining both in a unique pipeline with more accurate results than previous works and a faster (and constant) inference time (see \cref{fig:teaser_new}). We also show that such a combination of classification and retrieval is not effective with currently viable classification methods for planet-wise geo-localization due to their lack of accuracy.

\myparagraph{Contributions}
To summarize our contributions:
\begin{itemize}[noitemsep,topsep=1pt]
    \item We are the first to tackle the problem of fine-grained (error $\leq$ 25m) and city-wide (map area $>$100 km$^2$) VPR through classification, demonstrating the inadequacy of existing global-scale approaches in this setting and proposing a first feasible solution (D\&C).
    \item In D\&C we propose a new classifier, named Additive Angular Margin Classifier (AAMC), which uses the learned prototypes from a Additive Angular Margin Loss to classify new images. The AAMC is scalable and produces remarkably robust results.
    \item We show that our method not only achieves competitive results with retrieval methods, but above all it can be combined to create a fast, scalable and accurate pipeline that harnesses the best of both worlds.
\end{itemize}

Code and trained models will be made publicly available upon acceptance of this paper.

\section{Related Work}
\label{sec:related_work}

\myparagraph{Retrieval-based VPR}
Most commonly, approaches for VPR are retrieval-based, where the predicted coordinates are obtained with a similarity search over a database of pre-computed embeddings of either local \cite{Csurka_2003_bow, Jegou_2014, Jegou_2011_vlad, Perronnin_2010_fisher} or global features \cite{Arandjelovic_2018_netvlad,Kim_2017_crn,Liu_2019_sare,Zaffar_2020,Zaffar-2021}.
In recent works, deep feature-extractors have become the de facto standard, complemented with an aggregation or pooling layer. A notable example of the former is NetVLAD \cite{Arandjelovic_2018_netvlad}, in some cases equipped with attention mechanisms \cite{Kim_2017_crn}. Popular pooling strategies are \cite{Razavian-2015, Tolias-2016, Radenovic_2019_gem}.
A complete discussion on retrieval-based algorithms can be found in \cite{Berton_2022_benchmark,Masone_2021_survey}.
All these approaches leverage contrastive learning via triplet loss, which requires a mining procedure to compute a cache containing suitable triplets, to be updated periodically. 
Due to this formulation, all retrieval pipelines suffer from poor scalability: at \textit{training time}, as the database size increases, the computation required for mining becomes predominant with respect to the actual training; at \textit{test time}, the time required to perform the similarity search grows linearly with the database size, thus leading to potentially unacceptable delays for applications. 
Recently, CosPlace \cite{Berton_CVPR_2022_cosPlace} has addressed the training time scalability problem by using an alternative approach to contrastive learning, allowing to learn from large scale databases and achieving state of the art results across many datasets. Yet, it uses retrieval for inference, thus the scalability problem at test time still persists.
Concurrently to our work, in \cite{vallina_2023_gcl} the authors propose a general formulation of the contrastive loss that does not require hard pair mining.

\begin{figure}[t]
    \begin{center}
    \includegraphics[trim={2cm 2cm 2cm 2.5cm},clip,width=1\columnwidth]{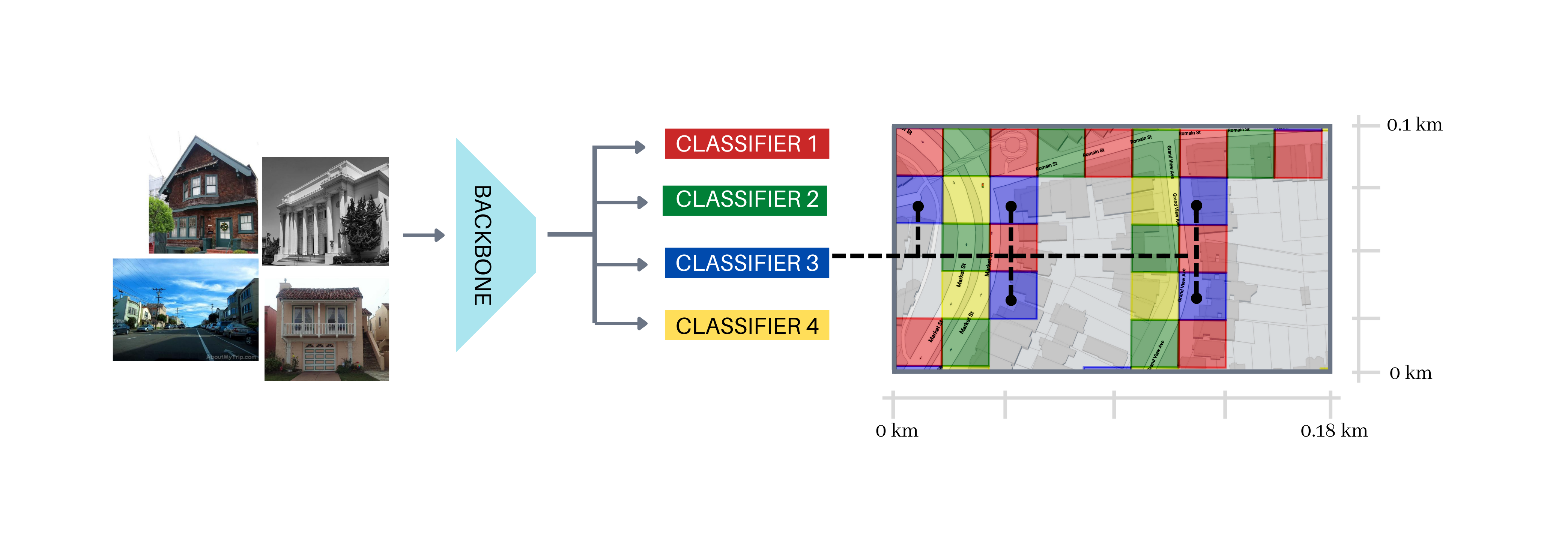}
    \end{center}
    \vspace{-0.2cm}
    \caption{\textbf{Architecture of {\our}}.The different groups, and their relative classifiers, are color coded. The picture explicates how cells are distributed into groups.
    }
    \label{fig:architecture}
\end{figure}

\myparagraph{Classification-based VPR}
There is another branch of literature that casts the place recognition problem as a classification task, but on a world-wide scale. 
Seminal works in this setting are RevIm2GPS \cite{Vo_2017_revIm2GPS} and PlaNet \cite{Weyand_2016_PlaNet}. These global-scale classification methods partition the earth into a set of geo-classes, and the geographic center of the predicted class is used as output geolocation.
This approach to VPR has many advantages, mainly in terms of space and time complexity. However, the final accuracy is highly dependent on the adopted partitioning scheme. Fine-grained partitions are necessary to increase the resolution of the localization, but scaling up the number of classes is not trivial as it causes the number of parameters to grow and reduces the number of samples per class.
In Hierarchical Geolocation Estimation (HGE) the authors exploit a geographic hierarchy of classes to mitigate this issues \cite{Muller_2018_hierarchical_geolocation},  whereas CPlaNet \cite{Seo_2018_CPlaNet} is based on a combinatorial partitioning of multiple geoclass sets. 
Other works \cite{Kordopatis_2021_EfficientNetGeoloc, Izbicki_2020_earth_geoloc} aim at learning the  classes centers.
Recently, \cite{Theiner_2022_WACV} proposed a partitioning based on interpretable hierarchies (e.g. city, region, country).
For the fine-grained urban setting, the idea of addressing it as a classification problem has been explored back in 2013 \cite{Gronat-2013} using SVM classifiers, however only in a small geographical area ($\approx 1.56$ km$^2$) and not on a city-wide scale.
In robotics, where VPR is often implemented as a part of the localization stack of a mobile robot, classification based approaches are often preferred to retrieval thanks to their lower requirements. 
Researchers in this field have focused on creating lightweight classification frameworks that can be executed with extremely low-latency on the limited onboard hardware of the robots~\cite{Chancan_2020_Flynet,Arcanjo_2022_Drosonet}.However, these methods are not built for accuracy on large scale problems.

\myparagraph{Relationship with prior works}
Our work is related to both retrieval and classification methods; we leverage intuitions from previous state-of-the-art (SOTA) in both fields and build a new technique specifically suited to tackle the city-wide localization task.

With respect to CosPlace~\cite{Berton_CVPR_2022_cosPlace} (\textbf{retrieval-based} SOTA), we use the same concept of \textbf{Groups} \ie a set of geoclass partitionings. In CosPlace, only a subset of groups is used, as the target is learning to extract features meaningful for localization. Instead, in this work we aim at learning a distribution that covers geographically the entire city, so we train on all the Groups.
Moreover, in CosPlace the learnt classifiers (one for each group) are discarded at inference time, and predictions are obtained via similarity search in a database, thus incurring in costly memory and time requirements. Contrarily, we keep all the classifiers and we exploit their \quotes{collective wisdom} to quickly obtain predictions.

Regarding \textbf{classification-based} methods, there is not a clear SOTA: the most popular works PlaNet \cite{Weyand_2016_PlaNet} and CPlaNet \cite{Seo_2018_CPlaNet} train on private datasets, thus resulting in unfair comparisons to other methods, and they have no public implementation.
Comparisons in previous literature are further hindered
by the fact that methods rely on different backbones. 
Therefore, ours is the first fair comparison of existing classification methods in a fine-grained and city-scale setting.
In terms of similarities with our method, CPlaNet also uses the idea of merging predictions from multiple classifiers: the authors create 5 discrete partitions of the earth, and perform inference  on the intersection of the classes in these partitions, with a combinatorial scheme. 
These partitions are overlapping and each one contains geographically adjacent classes. On the contrary, we show that in the geographically dense city-wide setting (as opposed to the sparse planet scale one) a discrete partitioning of adjacent classes seriously impedes the learning capabilities of a classifier and our partition is designed to prevent this phenomenon.
In \cref{sec:partitioning_method} we further detail this reasoning, and experimental results in \cref{sec:experiments} confirm this claim.
Another key difference is that CPlanet's classes are irregularly shaped, formed as intersection of overlapping partitions and logits are assigned using a combinatorial scheme. As a result, inference is costlier. 
Instead, our method uses a simple partitioning scheme that makes it faster, while also largely outperforming competitors.

Finally, we experimentally demonstrate in \cref{sec:classification_plus_retrieval} that our approach can be combined with retrieval methods into a single pipeline with faster inference time and better accuracy. This is possible thanks to D\&C's accuracy which allows to use it as a distractor-filter for the retrieval, whereas the existing classification-based approaches are too imprecise and combining them with retrieval worsens results.

\section{Method}
\label{sec:method}

\begin{figure}[t]
    \begin{center}
    \begin{minipage}{.0265\textwidth}
        \includegraphics[width=\textwidth]{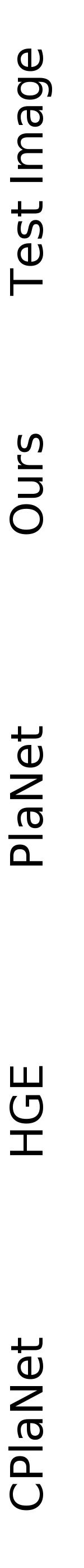}
    \end{minipage}
    \begin{minipage}{.445\textwidth}
        \includegraphics[width=\textwidth]{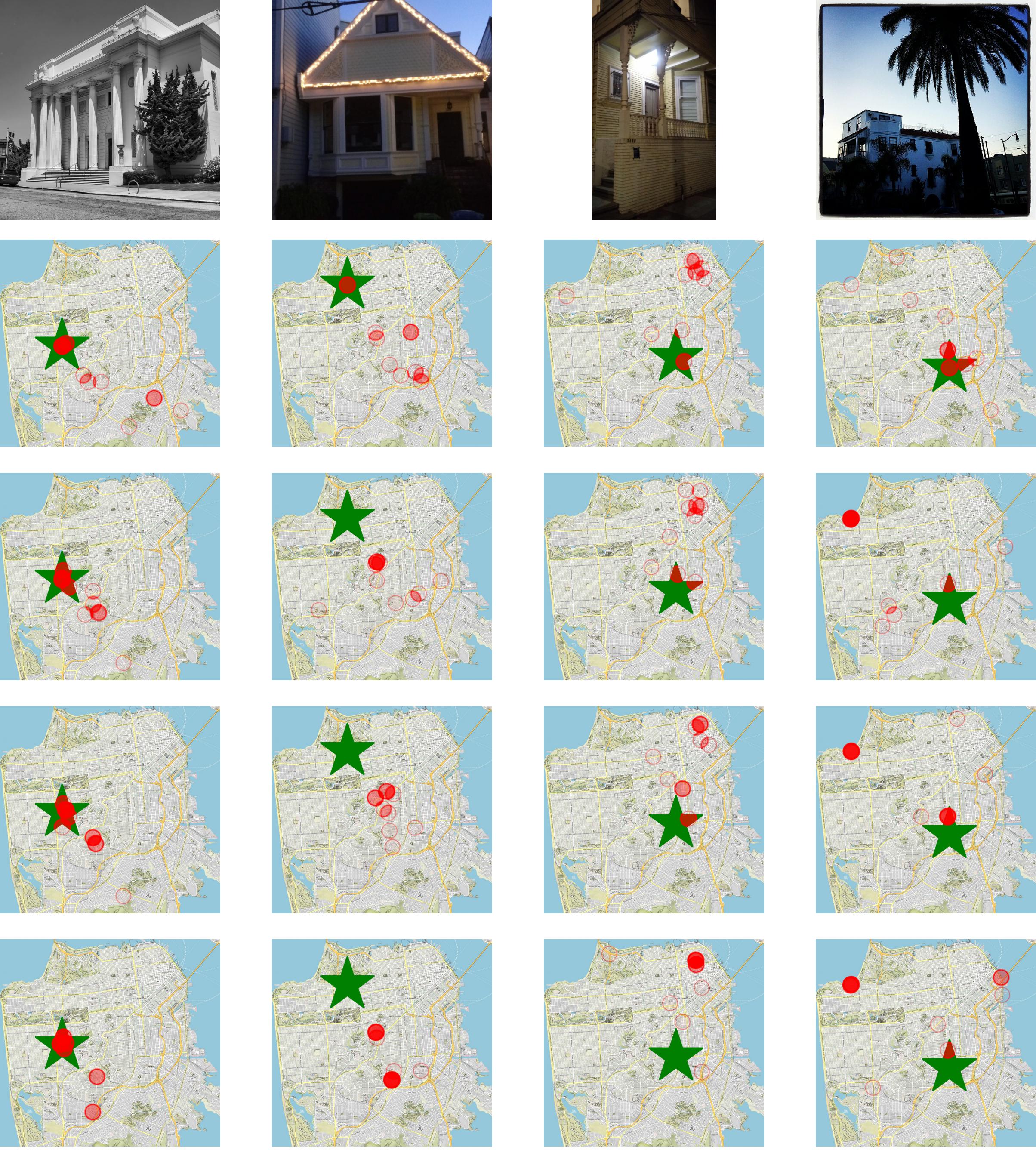}
    \end{minipage}
    \end{center}
\caption{\textbf{Qualitative examples of predictions for each method.} The green star shows the ground truth of each test image, while the red circles represent the first 10 predictions. Brighter red indicates multiple predictions close to each other.}
\label{fig:grid_preds}
\vspace{-0.5 cm}
\end{figure}

We consider the VPR task in a large urban environment (\ie a city covering $>$ 100km$^2$) and we deem a place correctly recognized if its predicted location is within 25m from the ground truth (as commonly done in literature \cite{Arandjelovic_2018_netvlad, Warburg_2020_msls, Liu_2019_sare, Ge_2020_sfrs, Berton_2021_svox, Hausler_2021_patch_netvlad, Wang_2022_TransVPR, Berton_2022_benchmark,Zaffar-2021}).
Furthermore, we assume to have a training set of geo-tagged urban images $\mathcal{T}=\{(I_i, \, east_i, \, north_i)\}$, where $I_i$ is an image and the pair $(east_i, \, north_i) : \mathbb{UTM}_e \times \mathbb{UTM}_n$ represents its UTM coordinates (which are an approximation of GPS coordinates of a local geographical area over a flat surface).
The training set $\mathcal{T}$ is created by sampling densely over the city (\ie roughly one image per meter of road), because we want to achieve precise localization. 

With this setting, our goal is to learn from $\mathcal{T}$ a classifier to perform the VPR task.
In pursuit of this objective, we rely on two key ingredients. Firstly, we partition $\mathcal{T}$ in disjoint sets of cells (\ie classes) so that there is no visual overlap among distinct classes included in the same split.
Secondly, we use an ensemble of classifiers, one for each split of $\mathcal{T}$. In particular, we propose a novel type of classifier, called Additive Angular Margin Classifier (AAMC), which inherits the discriminative power of the Additive Angular Margin Loss \cite{Wang_2018_cosFace}. 
We thoroughly explain our partitioning method in \cref{sec:partitioning_method} and the AAMC in \cref{sec:lcml}.

\subsection{Partitioning method}
\label{sec:partitioning_method}

Splitting the training set $\mathcal{T}$ in classes is non-trivial given that the label space (GPS or UTM coordinates) is continuous. 
Moreover, in light of the high, uniform density that we deal with, learning over a discretized geographical partitioning with a vanilla classification framework is an ill-posed problem. The reason for this is that images at the boundaries of neighboring cells can have very similar appearance, and consequently similar embeddings, while pertaining to different classes. This results in noisy gradients and overall it hinders a model's learning ability.
To address this issue, we adopt a partitioning method that produces multiple disjoint splits of the training set, preventing any classes in the same split from sharing a boundary.
In practice, this avoids \textit{perceptual aliasing} between different classes in the same split.

The core idea of our partitioning method is to \textit{build different splits made of \textbf{non-adjacent} geographic cells, and learn a classifier on each split} (see \cref{fig:architecture}).
This way each cell (akin to a class) is assigned to a exactly one classifier, while its neighboring cells are assigned to separate classifiers.

The scheme that we adopt to split the dataset into cells is straightforward and it is inspired by CosPlace~\cite{Berton_CVPR_2022_cosPlace}: we divide the map into equal-sized square cells, that are defined by a single hyper-parameter $M$, representing the length of the side of the square.
Each cell corresponds to a class. 
Since the geo-localization task requires to output a set of coordinates, we define a simple function Class2UTM that maps each class to a set of coordinates (the center of its corresponding cell), which will be used as prediction.
Formally, a class $C_{e_i, n_j} \in \mathcal{S}$ is defined as :
\begin{equation}
    \label{eq:classes}
    C_{e_i, n_j} = \left\{ (east, north) : \left\lfloor\frac{\textit{east}}{M} \right\rfloor = e_i,  \left\lfloor\frac{\textit{north}}{M}\right\rfloor = n_j \right\} 
\end{equation}
and the function Class2UTM is 
\begin{equation}
    \label{eq:class2utm}
    \begin{aligned}
        \textrm{Class2UTM}:& \mathcal{S} \rightarrow \mathbb{UTM}_e \times \mathbb{UTM}_n, \\ 
        C_{e_i, n_j}  \mapsto \big( (e_i & +0.5)\cdot M, \; (n_j + 0.5)\cdot M\big)
    \end{aligned}
\end{equation}

Finally, we partition the set of cells into separate Groups: we therefore set a separate hyperparameter $N$ and define each Group as
\begin{equation}
    \label{eq:groups}
    \scalebox{1}{$G_{u v} = \big\{ C_{e_i n_j} : (e_i \bmod N = u) \; \land \; (n_j \bmod N = v) \big\}$}
\end{equation}

The parameter $N$ defines the minimum distance (in number of cells) between two nearby classes within the same split.
Moreover the value of $N$ determines the number of groups $|G|$ that are created from the dataset, with $|G| = N^2$.
In \cref{sec:ablations} we empirically find the best values for the hyperparameters to be $M = 20$ meters and $N = 2$.
Note that, unlike the partitioning used by CosPlace~\cite{Berton_CVPR_2022_cosPlace}, we do not need the training images to be labeled with a heading angle. This makes our solution more widely applicable.

\subsection{Additive Angular Margin Classifier (AAMC)}
\label{sec:lcml}
Using the Groups formulation we effectively obtain a set of $|G|$ independent partitions of the cells, so we can use a \textit{Mixture-of-Experts} approach and assign a separate classifier to each partition (see \cref{fig:architecture}).
Recently, several studies \cite{Deng_2019_arcFace, Wang_2018_cosFace, Berton_CVPR_2022_cosPlace} have found that using large-margin losses when training siamese architectures for retrieval leads to more discriminative embeddings and a better structured latent space \cite{emb_struct1, emb_struct2}.
However, these approaches have only ever been used to train a feature extractor; thus the learnt prototypes were never applied to perform classification at test time.
Motivated by the fact that we want to enjoy the benefits that these losses provide, we propose a novel classifier that allows to exploit the highly informative prototypes learnt during training.

To this end, we build upon the Additive Angular Margin Loss (ArcFace) \cite{Deng_2019_arcFace}, which has been used in prior large scale retrieval works~\cite{Wang_2018_cosFace, Srivastava_2019_face_bench, Berton_CVPR_2022_cosPlace}.
The ArcFace allows to learn highly discriminative embeddings by maximizing the inter-class angular distance, measured via the cosine similarity between a matrix of learnable class prototypes and the normalized embeddings. Formally, the ArcFace loss is defined as
\begin{equation}
    \mathcal{L}_\textit{arc} = 
    \frac{1}{N} \sum_{i}{-\text{log} \frac{e^{s\cos (\theta_{y_i} + m)} }
         {e^{s\cos(\theta_{y_i} + m)} + \sum_{i \neq j}{e^{s\cos \theta_j}}}
    }
\end{equation}
subject to 
\begin{equation}
\begin{aligned}
    \cos \theta_j &= W_j^T x_i \\
    W = \frac{W^*}{||W^*||}, \quad &x = \frac{x^*}{||x^*||}
\end{aligned}
\end{equation} 
where $x_i$ is the $i$-th embedding corresponding to the ground-truth class of $y_i$ and $W_j$ is the prototype vector of the $j$-th class. 

Previous works~\cite{Wang_2018_cosFace, Srivastava_2019_face_bench, Berton_CVPR_2022_cosPlace} only use the ArcFace (or similar angular margin losses) to learn to extract embeddings that can be used in a retrieval pipeline, discarding the prototypes.
Instead, we argue that these prototypes learn a meaningful mapping from the embeddings to each class and can therefore be exploited directly to classify.
To the best of our knowledge, this is the first work that uses the prototypes learnt with the ArcFace at inference time.

We call this approach Additive Angular Margin Classifier (AAMC).
Namely, the generic  $k$-th classifier (\ie the classifier assigned to the $k$-th Group) is characterized by a matrix $W_k \in \mathbb{R}^{S_k \times d}$ of $d$-dimensional class prototypes, where $S_k$ is the number of classes in group $k$ and $d$ is the dimensionality of the backbone's embeddings. 
At inference time, given the embedding of a test image $x \in \mathbb{R}^{d} $, we extract a set of $|G|$ normalized predictions
$\{p_k = \text{softmax}(W_k^T x) \in \mathbb{R}^{S_k}\}_{k = 1}^{|G|} $.

Among them, we choose the one with maximum confidence across all classifiers
\begin{equation}
    \begin{aligned}
        \textrm{D\&C}:  \mathbb{R}^{S_1}  \times \ldots \times \mathbb{R}^{S_{|G|}}  \rightarrow &\mathbb{UTM}_e \times \mathbb{UTM}_n, \\
     \{p_k\}_{k=1}^{|G|} \mapsto  \textrm{Class2UTM} ( \underset{k \in 1..|G|}{\text{argmax}} &\; \underset{c \in S_k}{\text{argmax}}   \ p_k(c) ) 
    \end{aligned}
\end{equation}
where $p_k(c)$ denotes the $c$-th element of the vector $p_k$.
In summary, each classifier provides the logits distribution on the classes pertaining to its group. To obtain the final prediction on the entire geographical support for a given image, we choose the single prediction with the highest confidence across all AAMC classifiers.

The intuition behind our \textit{mixture-of-AAMC} approach is that while the class containing a given test image only belongs to a single classifier, the other classifiers are likely to predict nearby classes, given that classes close to each other share similar visual features.
In \cref{fig:std} we further detail this behavior, studying the agreement and the correlation among prototypes in different AAMC classifiers.


\section{Experiments}
\label{sec:experiments}

\begin{figure}
    \begin{center}
    \includegraphics[width=\columnwidth]{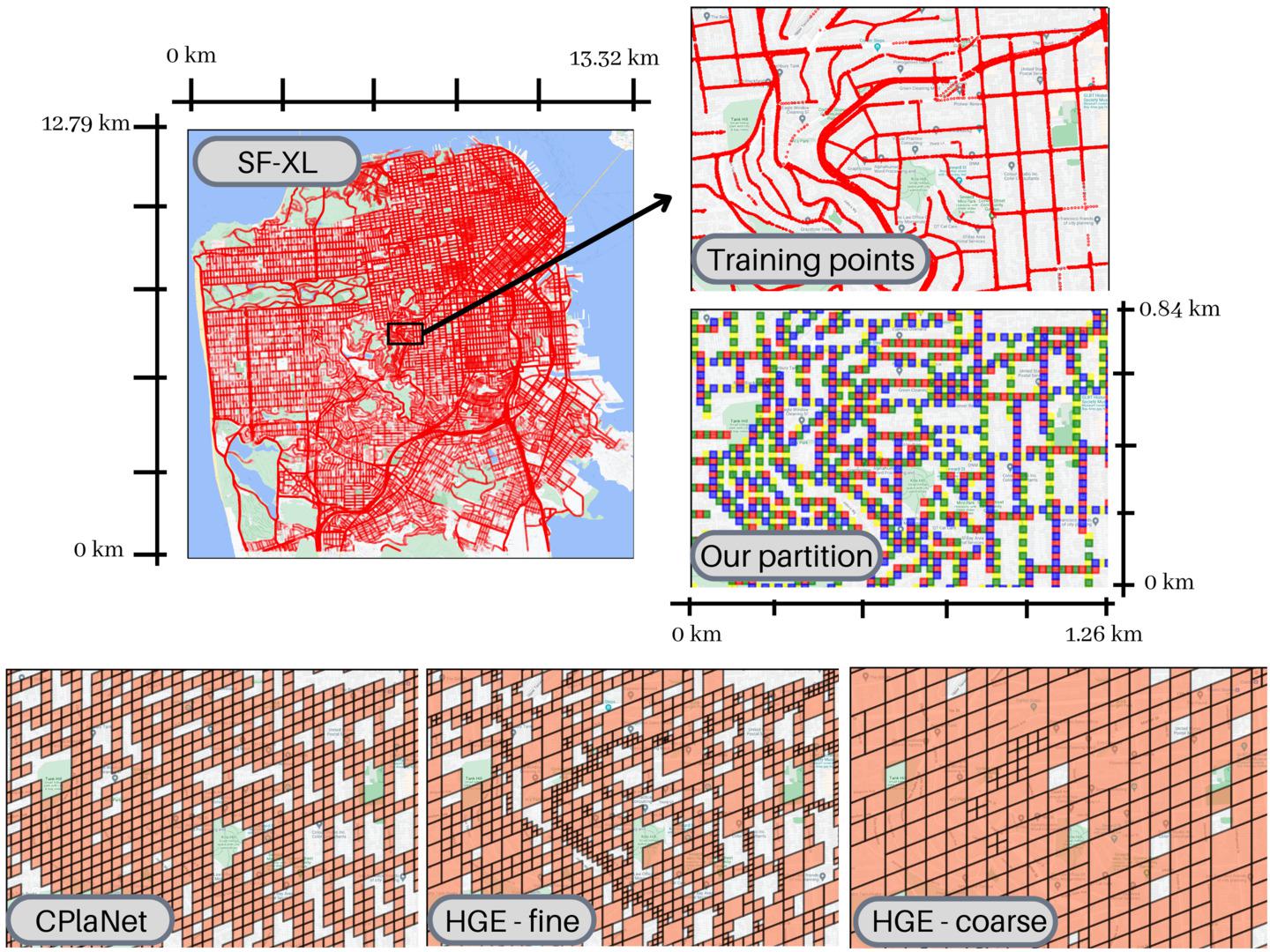}
    \end{center}
\caption{
\textbf{Maps showing multiple partition methods over the training data of SF-XL.}
}
\label{fig:sf_xl_partitions}
\end{figure}

\subsection{Experimental Setting}

\myparagraph{Implementation details}
\label{sec:implementation_details}
The split of the dataset in classes is performed using $M=20$, meaning that each class is a square cell of 20 meters per side.
The classes are then grouped together using $N=2$, which leads to the creation of 4 groups and 4 classifiers within our model.
This creates over 113k classes for the SF-XL training set; \cref{fig:sf_xl_partitions} gives a visual representation of the outcome. Further analysis on the number of classes generated over the various datasets can be found in the Supplementary.

The classifiers are trained independently one per epoch.
We train for 1M iterations with batch size 64.
Given the total of 4 classifiers, each ons is therefore trained once every 4 epochs using Adam \cite{Kingma_2014_adam} as optimizer with learning rate of $1e^{-4}$.
Following \cite{Kordopatis_2021_EfficientNetGeoloc},
we use an EfficientNet as backbone.
For fairness, we compute results using the same batch size, data augmentation, backbone and training set for all classification methods.
Note that this is in contrast with previous works \cite{Weyand_2016_PlaNet, Seo_2018_CPlaNet, Muller_2018_hierarchical_geolocation, Kordopatis_2021_EfficientNetGeoloc, Izbicki_2020_earth_geoloc} that use different backbones (and different training dataset) for all methods (see \cref{sec:related_work}), which makes it difficult to discern if the reported performance improvements was given by the method, a stronger backbone or better training data.
For D\&C, the backbone is followed by an average pooling and a whitening layer \cite{Radenovic_2019_gem}, which provides the input for the classifiers.

To assess the quality of different techniques on the task of city-wide visual place recognition, we run our experiments on the large-scale dataset of San Francisco eXtra Large \cite{Berton_CVPR_2022_cosPlace}, which consists of a training set of 41.2M images that densely maps the whole urban area of San Francisco ($>$100 km$^2$). 
The dataset provides two test sets, namely \emph{test set v1} and \emph{test set v2}, both of which are collections of photos taken with smartphones.
To this date, this is the only dataset well representative of a fine-grained and city-wide VPR application, that is our target setting. 
Nevertheless, in \cref{sec:limitations} we also discuss the use of {\our} with much smaller urban datasets, covering less than 3 km$^2$, to discuss its limited effectiveness for small scale problems.

\myparagraph{Metrics}
Visual place recognition methods are assessed using different metrics, depending on the type of method. Traditionally, retrieval methods for fine-grained VPR are evaluated using the recall@N (R@N) with a threshold of 25 meters \cite{Arandjelovic_2018_netvlad, Liu_2019_sare, Kim_2017_crn, Ge_2020_sfrs, Berton_CVPR_2022_cosPlace, Berton_2022_benchmark} for correct matches, whereas classification methods for planet-scale localization use the  Great Circle Distance (GCD) \cite{Hays_2008_im2gps}, which measures accuracy within a given threshold distance.
In order to properly compare retrieval and classification methods, we introduce a new metric called \textbf{Localization Radius@N} (LR@N) that summarizes both: \textit{Given the top-N predictions for a certain localization algorithm, it evaluates the percentage of queries which are correctly localized within 25 meters of the ground truth, in at least one of the top-N predictions.}

Although the LR@N is equivalent to the R@N @25m used in retrieval, its definition makes it compatible with a classification pipeline. In fact, the LR@1 is also equivalent to GCD@25m.

\myparagraph{Baselines overview} We split the baselines in three main categories:

1) Previous \textit{classification methods}, designed for world-wide geolocalization, that can be readily repurposed for city-wide geolocalization.
Among these, we use PlaNet \cite{Weyand_2016_PlaNet}, 
Hierarchical Geolocation Estimation (HGE) \cite{Muller_2018_hierarchical_geolocation},
CPlaNet \cite{Seo_2018_CPlaNet} and
MvMF \cite{Izbicki_2020_earth_geoloc}.
Note that, unlike {\our}, all these methods rely on the S2 Sphere library, and use hyperparameters for class partitions that have been tuned on global-scale classification: for fair comparison between methods, we performed a thorough grid search of such hyperparameters on SF-XL, which we used to compute the results in \cref{tab:main_table}. The final splits for each method are reported in \cref{fig:sf_xl_partitions}.
To compute results with classification methods (some of which do not have open source implementations), we reproduced their results on world-wide datasets, and then we used the code to train on the city-scale setting. We will release the code also for the implementations of such methods like PlaNet, CPlaNet, and MvMF.
We provide further analysis on the optimal hyperparameters for each partitioning method in the Supplementary.

2) \textit{Retrieval methods} for visual place recognition, using deep neural networks to produce descriptors which are then matched to the query's.
For our experiments we use the ever-green NetVLAD \cite{Arandjelovic_2018_netvlad},
 CRN \cite{Kim_2017_crn}, SARE \cite{Liu_2019_sare}, SFRS \cite{Ge_2020_sfrs} and CosPlace \cite{Berton_CVPR_2022_cosPlace}.

3) \textit{Retrieval methods} combined with \textit{Approximate Nearest Neighbor} (ANN),
in a pipeline that uses the best retrieval method, namely CosPlace, with the ANNs that have shown best results, namely Hierarchical Navigable Small Worlds (HNSW) \cite{Malkov_2020_hnsw} and Inverted File Index with Product Quantization \cite{Sivic_2003_ivf, Jegou_2011_productQ}.
We chose the two optimal configurations, one designed for speed or LR@1.
Extensive experiments with other ANN methods, as well as different hyperparameters, are shown in the Supplementary.

\begin{table}
\begin{center}
\begin{adjustbox}{width=\linewidth}
\begin{tabular}{lccccccccccccccccccccc}
\toprule
\multirow{2}{*}{Method} & 
Infer. & \multicolumn{2}{c}{LR@1} \\
\cline{3-4}
 & time & test v1 & test v2  \\
\hline
\emph{Classification} &  & \\
~~PlaNet \cite{Seo_2018_CPlaNet}                 & 12 ms & 24.5 & 53.1 \\
~~HGE \cite{Muller_2018_hierarchical_geolocation}& 15 ms & 27.0 & 56.4 \\
~~CPlaNet \cite{Seo_2018_CPlaNet}                & 17 ms & 27.4 & 64.1 \\
~~MvMF \cite{Izbicki_2020_earth_geoloc}          & 12 ms & 22.6 & 52.2\\
\hline
~~\textbf{\ours (ours)}                          & 12 ms & \underline{61.0} & \underline{79.1} \\
\hline
\hline
\emph{Retrieval} &  & \\
~~NetVLAD \cite{Arandjelovic_2018_netvlad}    & 12117 ms & 40.0 & 71.1 \\
~~CRN \cite{Kim_2017_crn}                     & 12117 ms & 45.8 & 76.4 \\
~~SARE \cite{Liu_2019_sare}                   & 12117 ms & 45.5 & 78.8 \\
~~SFRS \cite{Ge_2020_sfrs}                    & 12117 ms & 51.2 & 83.1 \\
~~GeM \cite{Radenovic_2019_gem}               &  1514 ms & 21.7 & 43.1 \\
~~CosPlace \cite{Berton_CVPR_2022_cosPlace}   &  1514 ms & \underline{64.7} & \underline{83.4} \\
\hline
\multicolumn{2}{l}{\emph{Retrieval + Approximate Nearest Neighbor}}   & \\
~~CosPlace \cite{Berton_CVPR_2022_cosPlace} + HNSW \cite{Malkov_2020_hnsw}      &  4 ms & 52.5 & 77.8 \\
~~CosPlace \cite{Berton_CVPR_2022_cosPlace} + IVFPQ \cite{Jegou_2011_productQ} *&  8 ms & 55.1 & 82.6 \\
~~CosPlace \cite{Berton_CVPR_2022_cosPlace} + IVFPQ \cite{Jegou_2011_productQ} *& 141 ms & \underline{63.7} & \underline{83.3} \\
\hline
\emph{Mixed pipeline} &  & \\
~~\textbf{\ours (ours)} + CosPlace \cite{Berton_CVPR_2022_cosPlace} & 30 ms & \textbf{71.4} & \textbf{87.6} \\
\bottomrule
\end{tabular}
\end{adjustbox}
\end{center}
\caption{\textbf{Comparison of results for a large number of methods using different approaches.} All inference times measures are averaged over 1000 queries, on a system with a RTX 3090 GPU and i9-10940X CPU. The FAISS library \cite{Johnson_2019_faiss} is used for all nearest neighbor implementations.
\textit{Mixed pipeline} is the best configuration from \cref{tab:classification_plus_retrieval}, which performs retrieval on the top-100 classes obtained through D\&C.
*We show two versions of the Inverted File Index with Product Quantization, one tuned for speed and one for recall.
}
\label{tab:main_table}
\vspace{-0.4cm}
\end{table}

\subsection{Main Results}
\myparagraph{Comparison with previous works}
We report quantitative comparisons between {\ours} and existing baselines in \cref{tab:main_table}.
The results show some clear trends, which can be summarized in a few points:
\begin{itemize}[noitemsep,topsep=1pt]
    \item Previous \textbf{classification methods}, being designed for the uneven distribution of world-wide datasets, fail to achieve competitive results on SF-XL. Qualitative results are reported in \cref{fig:grid_preds}.
    \item D\&C, being specifically designed for learning in a dense urban setting is able to outperform previous classification methods on city-wide geolocalization, and it is almost competitive with the retrieval-based state of the art, (LR@1 of 3.7 points lower on the \textit{test v1}).
    \item \textbf{Approximate Nearest Neighbor} (ANN) search algorithms provide different implementations, some of which are able to speed up retrieval methods by 10x with a slight drop in LR@1, and others speeding up retrieval by almost 400x at the price of a 12 points drop in LR@1 on the \textit{test v1}. 
    \item The \textbf{Mixed Pipeline} of SOTA classification (D\&C) and retrieval (CosPlace) methods reaches speed on pair with classification methods, while providing a large improvement over any other results. More details on this mixed pipeline are presented in the next section.
    
\end{itemize}

\begin{table}
\begin{center}
\begin{adjustbox}{width=\linewidth}
\begin{tabular}{lcccccccc}
\toprule
&&&\multicolumn{3}{c}{Classification Method \& Time} \\
\cline{1-6}
\multirow{3}{*}{\begin{tabular}[c]{@{}c@{}}Retrieval\\Method\end{tabular}} &
\multirow{3}{*}{Top-N} & 
\multirow{3}{*}{\begin{tabular}[c]{@{}c@{}}kNN\\Time (\emph{ms})\end{tabular}} & 
D\&C  & HGE & CPlanet \\
& & & (12 ms) & (15 ms) & (17 ms) \\
\cline{4-6}
& & & LR@1 & LR@1 & LR@1  \\
\hline
\multirow{5}{*}{NetVLAD} & All &  12117 & 40.0 & 40.0  &  \underline{40.0} \\
  & 1000       & 42   & 50.6  & \underline{42.8}  & 38.7  \\
  & 100        & 4    & 56.7  & 41.1  & 37.4  \\
  & 10         & 0.6  & \textbf{62.6}  & 35.6  & 34.6  \\
  & 1          & 0.06 & 56.1  & 24.8  & 26.3  \\

\hline
\hline
\multirow{5}{*}{CosPlace} & All &  1514 & 65.9 & \underline{65.9}  & \underline{65.9} \\
& 1000       &8     & 70.3  & 62.0  & 57.5  \\
& 100        &1     & \textbf{71.4}  & 52.7  & 51.0  \\
& 10         & 0.1  & 70.2  & 40.9  & 42.4  \\
& 1          & 0.03 & 57.0  & 25.0  & 26.7  \\
\bottomrule
\end{tabular}
\end{adjustbox}
\end{center}
\caption{\textbf{Results of classification + retrieval pipelines} on SF-XL test v1.
The \emph{Top-N} column represents the number of cells within which we compute retrieval.
The rows with \emph{Top-N: All} are equivalent to using retrieval only, while the other rows employ the classification filter, reducing the search space by orders of magnitude. For retrieval, we use a VGG16 backbone. NetVLAD's dimensionality is 4096-D PCA (extraction time 2.1 ms), while CosPlace has 512-D (extraction time 5.1 ms).
}
\label{tab:classification_plus_retrieval}
\vspace{-0.4cm}
\end{table}

\subsubsection{Classification + retrieval}
\label{sec:classification_plus_retrieval}
\begin{figure}
    \begin{center}
    \includegraphics[width=0.7\columnwidth]{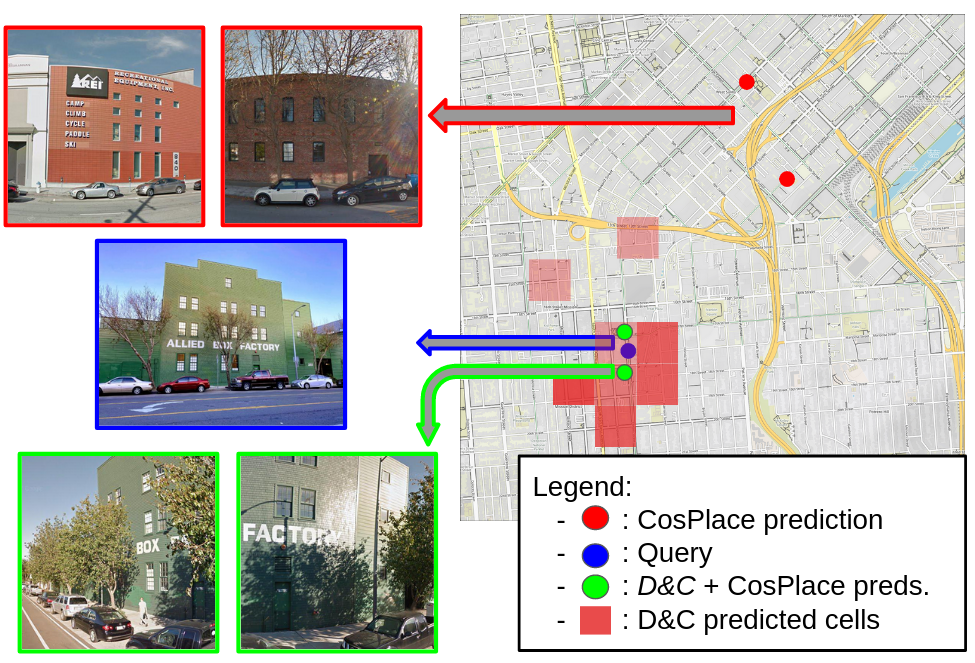}
    \end{center}
    \vspace{-0.3cm}
    \caption{Example of our \textbf{mixed pipeline}. Thanks to the reduction in the search space obtained via the predictions of \ours, the retrieval module correctly localizes the query }
    \label{fig:mixed_pipe}
    \vspace{-0.3cm}
\end{figure}

In this section we analyze how classification methods can be pipelined with retrieval ones into a single system, with the aim of improving accuracy and speed of results.
The idea is to restrict the search space for the retrieval's kNN search only to the cells that have been predicted with higher confidence by the classification model.
For example, when using only the first 10 classes (Top-10), images within the 10 cells where the model predicts the location of the test image with the highest confidence are then used for retrieval.
\Cref{fig:mixed_pipe} visually exemplifies how the filtering on the Top-N classes effectively removes distractors from the search space.
To provide a relevant analysis, we adopt different classifications models, namely HGE, CPlaNet and D\&C, while for retrieval we use models trained with NetVLAD \cite{Arandjelovic_2018_netvlad} and CosPlace \cite{Berton_CVPR_2022_cosPlace}, which are respectively the most popular and the most recent SOTA.
%
We report the results in \cref{tab:classification_plus_retrieval}: 
we separately show the multiple components of the total inference time using a two-stages pipeline, namely the classification, the descriptors extraction and kNN.
Regarding descriptors extraction, we only consider the extraction of the test image (query) descriptors, given that the ones from the database can be extracted offline. In the table, when Top-N=All it means that there is no filtering and the pipeline is the same as the pure retrieval method.

The results clearly show the huge benefit of using a two-stages classification + retrieval pipeline for large-scale visual geo-localization.
We find that such a pipeline leads to faster inference and better results \wrt a standard retrieval system.
The former is mostly due to the reduction in search space, which reduces the number of database descriptors by the kNN.
The latter is due to the filtering of cells to which the classification model assigns a low probability of containing the test image, therefore simplifying the retrieval task by eliminating distractors. 

Among the most notable results, using the Top-100 predictions from \ours, CosPlace achieves a new SOTA (+ 6\% LR@1), while being 500 times faster than the retrieval only version.
Similarly, the NetVLAD model can achieve a speedup by 4 orders of magnitude and an increase in LR@1 by 20\% when the retrieval is performed only on the Top-10 predicted classes.
\Cref{fig:teaser_new} visually shows how D\&C provides a much more scalable alternative to SOTA retrieval methods, even when retrieval is sped up by the best approximate nearest neighbor search.

\subsection{Ablations}
\label{sec:ablations}

\begin{figure}
    \begin{center}
    \includegraphics[width=0.65\columnwidth]{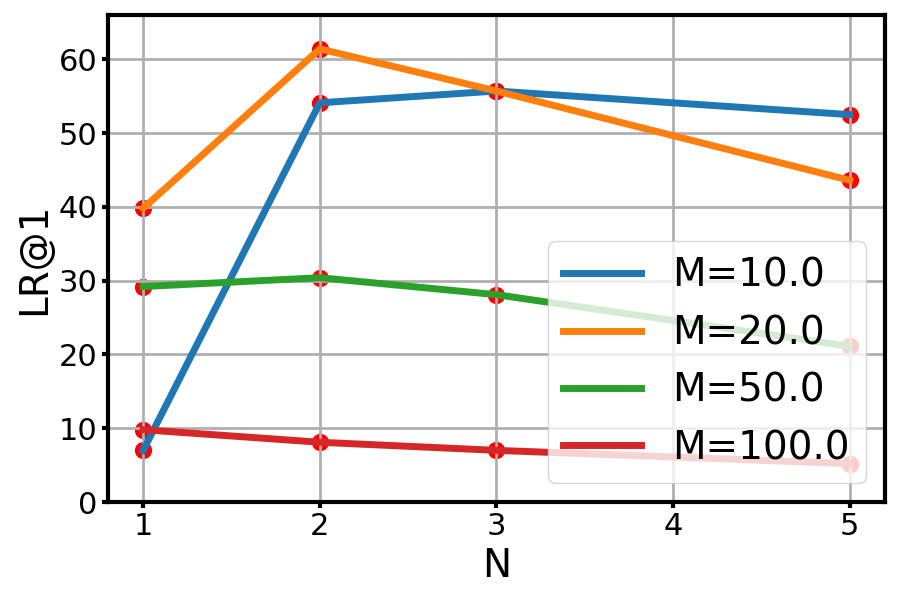}
    \end{center}
    \vspace{-0.3cm}
    \caption{\textbf{Ablation} on the values of $M$ and $N$. $M$ determines cell size, $N$ is the distance between cells in a group.}
    \label{fig:ablation_m_n}
    \vspace{-0.3cm}
\end{figure}

\myparagraph{Ablation on class partitions}
\Cref{fig:ablation_m_n} reports an ablation study on the values of $M$ and $N$.
Considering that the number of classifiers is $|G|=N^2$ (\cref{sec:partitioning_method}), 
each one trained independently once every $|G|$ epochs ( \cref{sec:implementation_details}), we can see that having a large number of classifiers leads to each one of them being trained too seldom for it to reach good performances;
on the other hand when using $N=1$ we incur in the learnability issues discussed in Sec. \ref{sec:partitioning_method}, due to adjacency among different classes of the same group.
Finally, $N=2$ stands out as the obvious best choice.

Regarding the value of $M$ (defining the side of a cell), $20$ m turns out to be the best choice. Values of $50$ m and $100$ m produce worse results. Understandably, this is because their classes encapsulate a greater variability wrt their finer counterparts, and thus are harder to learn.

\begin{table}
\begin{center}
\begin{adjustbox}{width=0.8\linewidth}
\centering
\begin{tabular}{lccccc}
\toprule
\multirow{2}{*}{Classifier} & 
\multicolumn{4}{c}{LR@N (at 25 m)} \\
\cline{2-5}
\multicolumn{1}{c}{}
& LR@1 & LR@5 & LR@10 & LR@20 \\
\hline
Cross-Entropy & 48.2 & 62.6 & 68.0 & 72.0 \\
AAMC          & \textbf{61.4} &  \textbf{73.6} &  \textbf{77.1} &  \textbf{79.6} \\
\bottomrule
\end{tabular}
\end{adjustbox}
\end{center}
\caption{\textbf{Ablation AAMC vs Cross-Entropy classifier.} This table clearly presents the benefit of our AAMC classifiers, which largely outperform standard linear classifiers trained with a cross-entropy loss.}
\label{tab:ablation_cosface_vs_celoss}
\end{table}

\myparagraph{Ablation on the loss} \label{par:loss_tsne} Once the dataset is split into groups and classes, a natural choice of loss would be the Cross-Entropy (CE) loss.
This can be easily implemented by using one linear layer for each group, training each classifier sequentially group by group.
However we empirically found (see \cref{tab:ablation_cosface_vs_celoss}) that the AAMC constantly outperforms a set of linear classifiers trained with a cross-entropy loss.
This is thanks to the formulation of the loss as margin maximization problem, that not only asks for classes separation (like in the vanilla CE), but also pushes them further away up to a margin.
This results in a better-structured feature space, as shown in the Supplementary where we analyze the t-SNE of the learnt embeddings.

\begin{figure}
    \begin{center}
    \begin{minipage}{.22\textwidth}
        \includegraphics[width=\textwidth]{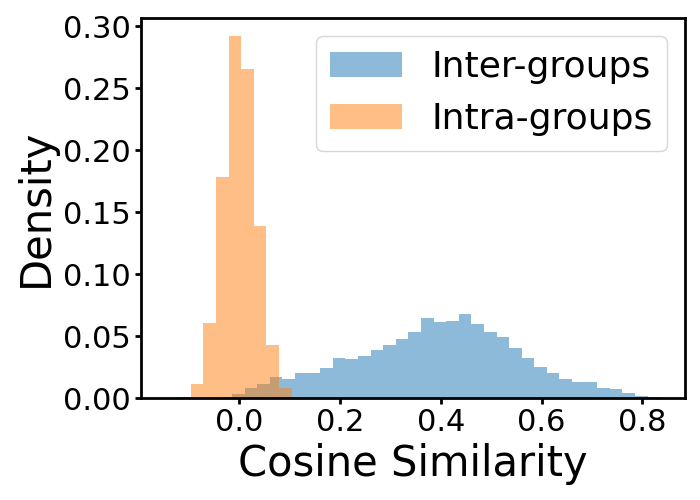}
    \end{minipage}
    \begin{minipage}{.22\textwidth}
        \includegraphics[width=\textwidth]{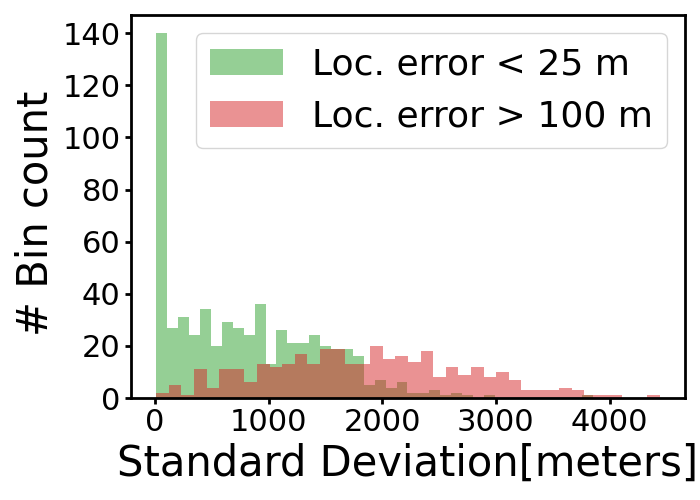}
    \end{minipage}
    \end{center}
    \caption{\textbf{Coordination of prototypes across groups.}
    (a) The left plot samples 500 neighboring prototypes (across all 4 groups), and shows their inter- and intra-group cosine similarity. It shows high correlation among inter-groups neighboring prototypes. Prototypes within a single classifier (intra-group) are well separated.
    (b) In the right plot we study the standard deviation (STD) among the prediction of each classifier.
    We can see that when the $N^2=4$ predictions are close to each other, the localization error is likely to be low ($<$ 25 meters), proving that the STD between D\&C's predictions from each expert is a good confidence measure, which is a very important value in real-world applications.}
    \label{fig:std}
    \vspace{-0.4cm}
\end{figure}

\myparagraph{Behavior of multiple classifiers}
Our method employs distinct AAMC classifiers, learned on disjoint sets, and merges their prediction to obtain the final logits. Seeing that prototypes of adjacent classes are learnt by different classifiers, it raises the question of "if and how they are related". A desirable property would be that prototypes become geographically correlated, on account of the fact that we want the different classifiers predictions to be consistent with one another.
In \cref{fig:std} we study this aspect.
In particular, the leftmost image samples 500 neighboring prototypes evenly distributed across groups, and plots the inter- and intra-group similarity distributions. The evidence indicates that the desired behavior is verified in practice.
The explanation for this self-emerging property is the same that motivates our use of non-adjacent class partitioning: images from neighboring cells share many visual features, which would confuse a single classifier trying to discern among them. Instead, in our partition, prototypes assigned to neighboring cells can easily fit their distribution, given that they are learnt independently from one another in their respective groups. At the same time, since these cells are similar, their prototypes end up being similar as well; and in the end the result is that prototypes are geographically correlated providing more robust predictions. 

Furthermore, having several classifiers predicting on disjoint sets also raises the question of what is their behaviour on samples that technically do not belong to any of their classes.
In \cref{fig:std} (right) we examine this aspect by studying the distribution of the Standard Deviation (STD) of the coordinates predicted by each classifier. Interestingly, the histogram clearly shows that when our system is able to precisely localize a sample (\textless 25m), all the classifiers concentrate their probability mass around the same area. On the contrary, on wrongly classified samples, the distribution reflects the uncertainty of the prediction. This indicates that classifier's agreement represents a good proxy of prediction reliability, which is an important feature in real-world applications.

\myparagraph{Further ablations}
More experiments on AAMC hyperparameters, learning curves and embedding visualizations can be found in our Supplementary Material.
\subsection{Limitations}
\label{sec:limitations}
{\our} is tailored for applications that aim at localizing in relatively large areas (e.g. SF-XL covers a surface of $170 ~km^2$) mapped with a dense training set.
We find that with smaller and less dense datasets like Pitts30k \cite{Torii-2015} and Tokyo 24/7 \cite{Torii_2018_tokyo247} (both are smaller than 3 km$^2$ and their density is less than half \wrt SF-XL)
retrieval methods are able to achieve very high recalls (over 80\% of recall@1 \cite{Ge_2020_sfrs, Berton_CVPR_2022_cosPlace},
whereas classification methods fail to achieve acceptable results (LR@1 lower than 50\%, regardless of the method).
This can be explained by the fact that for a given query, a single positive matching image in the database is enough for retrieval (the model matches the most similar image to the query),
whereas classification methods need a larger number of "positives", \ie an inadequate number of images for any given class leads to poor performances.
Thorough empirical experiments confirming this limitation of classification methods are shown in the Supplementary.

\section{Conclusions}
\label{sec:conclusions}

In this work we show the potential of framing the fine-grained VPR task in urban environments as a classification problem. 
We are the first to address this challenging scenario, proving that it is possible to achieve fine grained localization while obtaining a reliable measure of confidence in the predictions. 
We propose a novel inference pipeline to leverage the collective knowledge of a set of learnt classifiers that outperforms all the existing classification-based methods for localization.
Finally, we show how our proposed framework can be combined with retrieval methods obtaining an ideal trade-off between inference cost and localization performance, paving the way for faster and more accurate VPR systems.

{\small
\bibliographystyle{ieee_fullname}
\bibliography{paper_final}
}

\appendix

\section{Experiments}
In this Supplementary Material we report details that could not fit in the main paper.
In \cref{sec:further_ablations} we provide further ablations to better understand how our proposed method functions.

In \cref{sec:further_implementation_details} we provide a thorough discussion into how we adapted the partitioning scheme of previous works, that originally targeted planet-scale localization, for the proposed task of city-wide localization.

\subsection{Further Ablations}
\label{sec:further_ablations}
 
\myparagraph{Embedding learnt with our AMCC vs Cross Entropy}

\begin{figure}
    \centering
    \includegraphics[width=0.8\columnwidth]{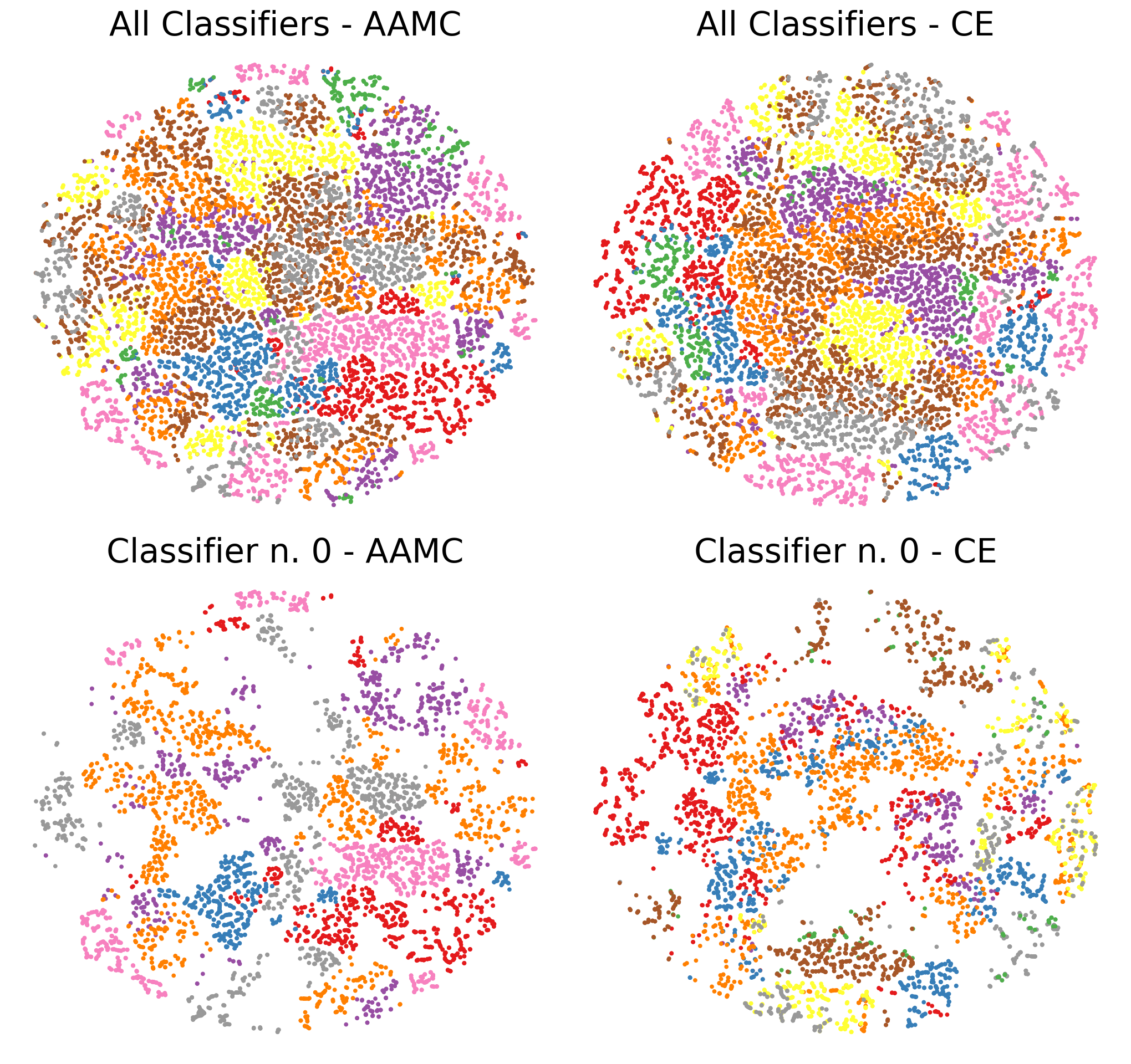}
    \caption{\textbf{t-SNE analysis} of embeddings in a 100m x 100m square. Each color codifies a different 20m cell.}
    \label{fig:tsne}
    \vspace{-0.2cm}
\end{figure}

The first row in \cref{fig:tsne} reports the t-SNE of all embeddings in a 100m square, with a model trained either with AAMC or a fully connected layer with cross-entropy loss; each color codifies a different 20m cell.
Even though some structures are visible, there is an amount of overlap which is understandable given that adjacent cells at such fine resolution can present high appearance similarities. The second row shows why in {\ours} each classifier is able to learn a meaningful distribution: inside each group, thanks to the non-adjacency of cells, classes are well defined. In particular, the two plots show how the AAMC yields a better-structured embedding space thanks to the concept of large margin.

\myparagraph{Behaviour of LR during training with different methods}

\begin{figure}
    \centering
    \includegraphics[width=0.73\columnwidth]{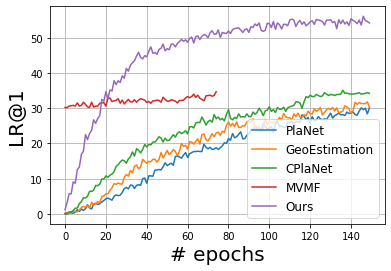}
    \caption{\textbf{Behaviour of LR@1 during training} for each of the methods. Note that MVMF \cite{Kordopatis_2021_EfficientNetGeoloc} starts with a high LR because it uses the weights of a trained PlaNet model.}
    \label{fig:supp_gcd_per_epoch}
\end{figure}

In \cref{fig:supp_gcd_per_epoch} we analyze how the LR@1 changes after each epoch for different methods (given the huge size of the dataset we define an epoch as 2k iterations).
We find that most previous works, namely PlaNet \cite{Weyand_2016_PlaNet}, CPlaNet \cite{Seo_2018_CPlaNet} and Hierarchical Geolocation Estimation \cite{Muller_2018_hierarchical_geolocation} present very mild improvements on LR within the first few epochs \wrt our D\&C, which on the other hand grows very steeply right from the beginning.
MvMF initializes its mixture assignment weights from a pretrained PlaNet model, and it terminates the training after less than 100 epochs.

\begin{figure}
    \centering
    \begin{minipage}{.23\textwidth}
        \includegraphics[width=\textwidth]{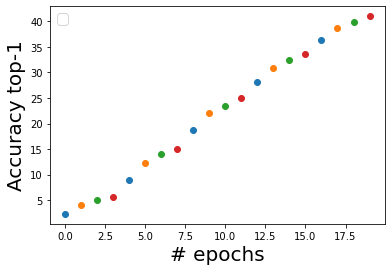}
    \end{minipage}
    \begin{minipage}{.23\textwidth}
        \includegraphics[width=\textwidth]{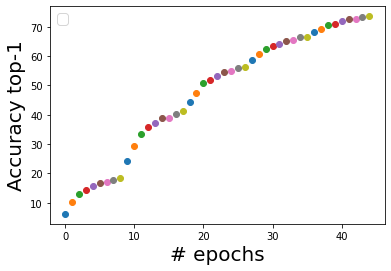}
    \end{minipage}
    \caption{\textbf{Evolution of classification accuracy during training with different values of N.} We can see that in the first epochs of training, the accuracy on the train set presents waves with period length of size $|G|$.
    Each color represents a different classifier being trained at the given epoch for a total of $|G|$ colors.
    }
    \label{fig:gcd_per_epoch_per_N}
    \vspace{-0.1cm}
\end{figure}

\myparagraph{Behaviour of classification accuracy during training using different $N$}
To better understand how the value of $N$ affects training stability, we built a plot using $N=2$ and $N=3$ and showing the accuracy on the train set at the end of each epoch.
The plot (\cref{fig:gcd_per_epoch_per_N}) shows that in the first epochs of training the accuracy forms waves with a period length of size $|G|=N \times N$, where $|G|$ represents the number of groups and the number of classifiers.
This is due to the fact that each classifier is trained once every $|G|$ epochs, meaning that at the $|G|th$ epoch the model will for the first time reuse a classifier that has been previously trained, resulting in a steep increase in accuracy every $|G|$ epochs.

\myparagraph{Ablation on AAMC hyperparameters}

Our AAMC classifier shares the 2 hyperparameter of 
the ArcFace formulation. Namely, \textit{s} determines the radius of the hypersphere onto which prototypes are projected, and \textit{m} is the enforced margin (in cosine space) between different prototypes.
\begin{figure}
    \begin{center}
    \includegraphics[width=0.23\textwidth]{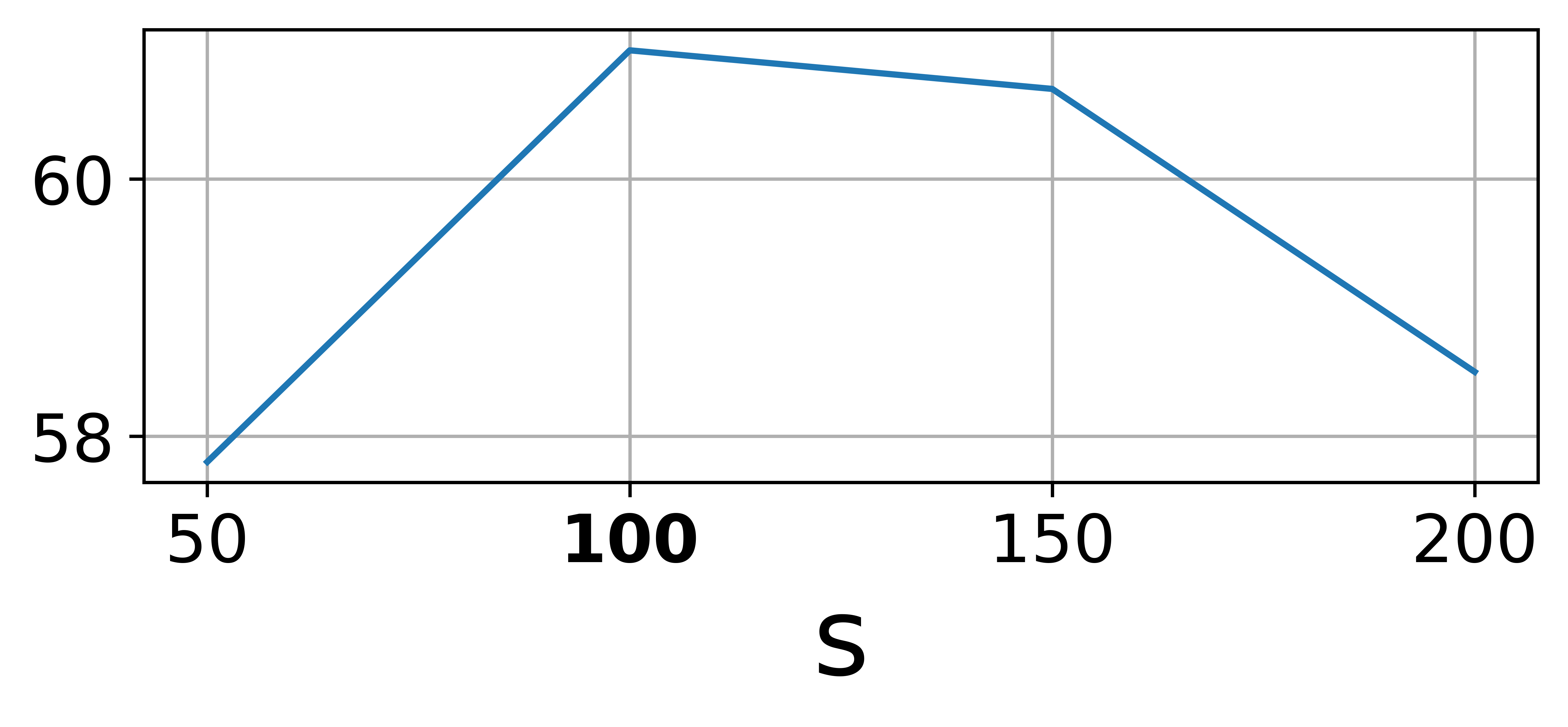}
    \includegraphics[width=0.23\textwidth]{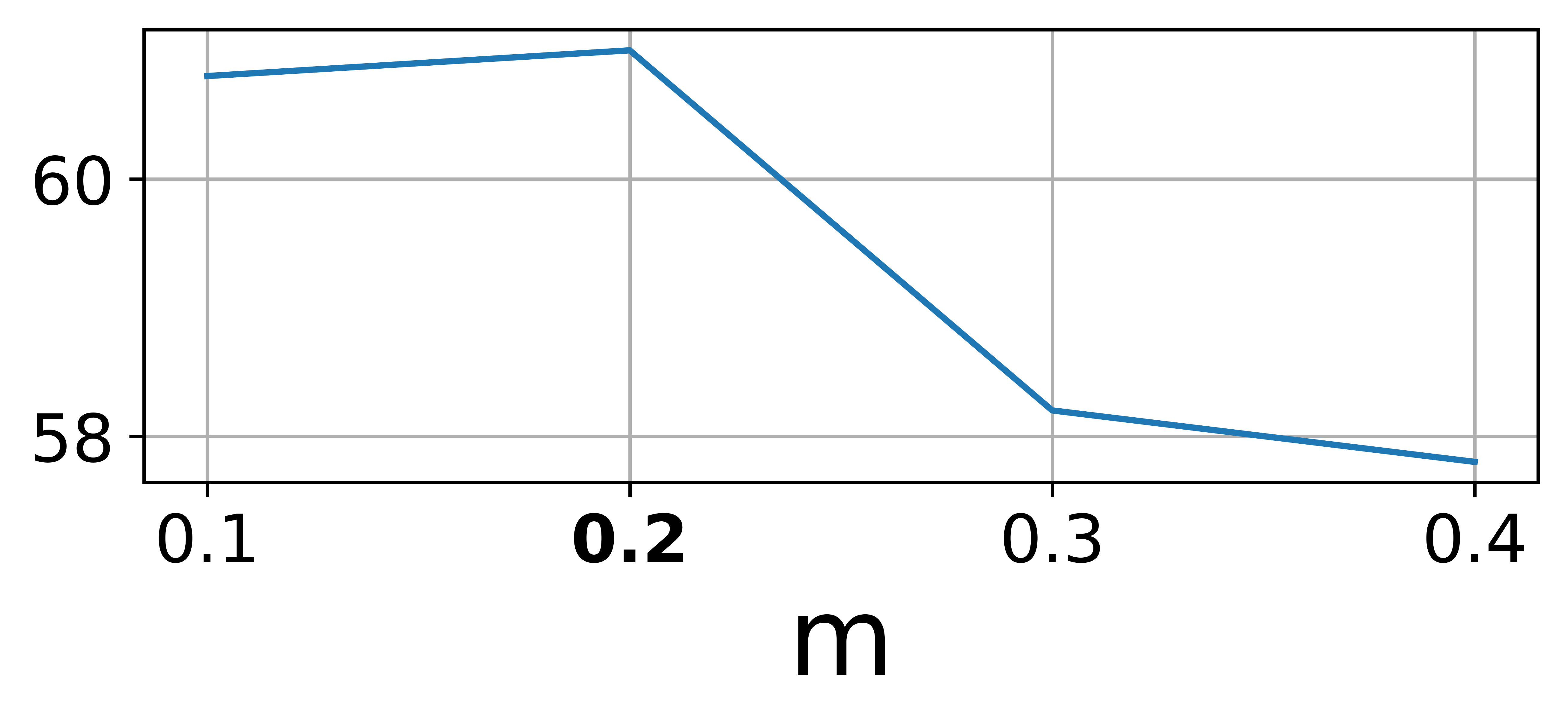}
    \end{center}
    \caption{
    \textbf{Ablation on \textit{s} (left) and \textit{m} (right).} Best values are highlighted in bold.}
    \label{fig:ablation_s_m}
\end{figure}

\begin{figure}[h!]    
    \begin{subfigure}{\textwidth}
    \includegraphics[width=0.5\columnwidth]{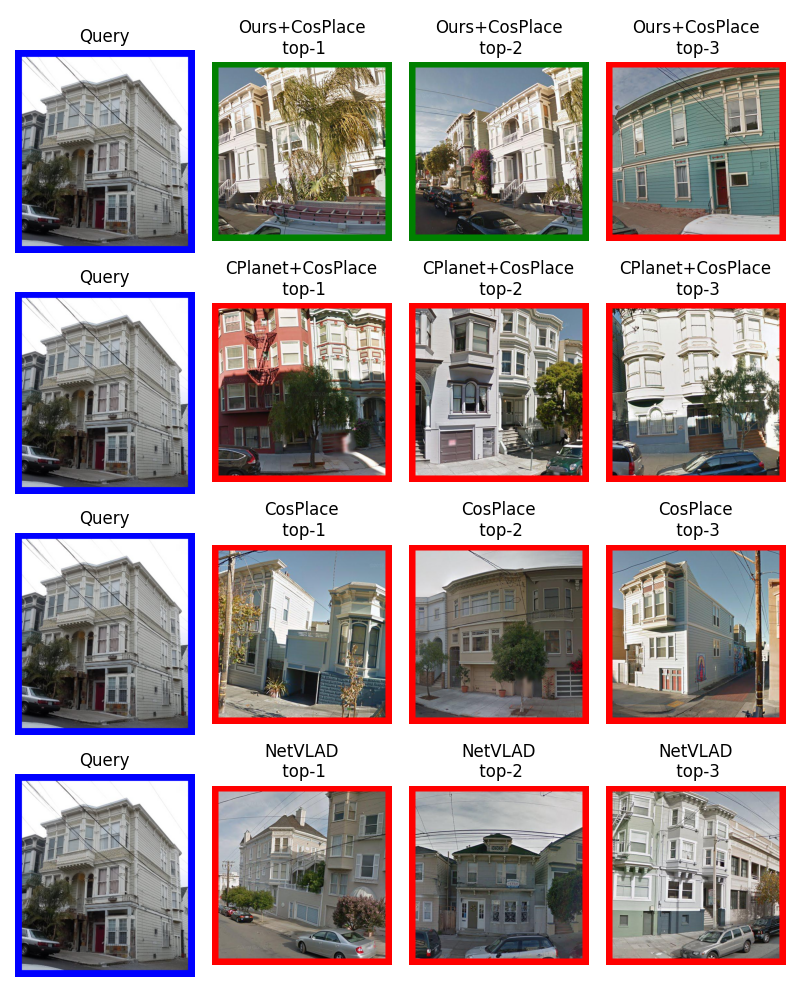}
    \end{subfigure}
    \begin{subfigure}{\textwidth}
    \includegraphics[width=0.5\columnwidth]{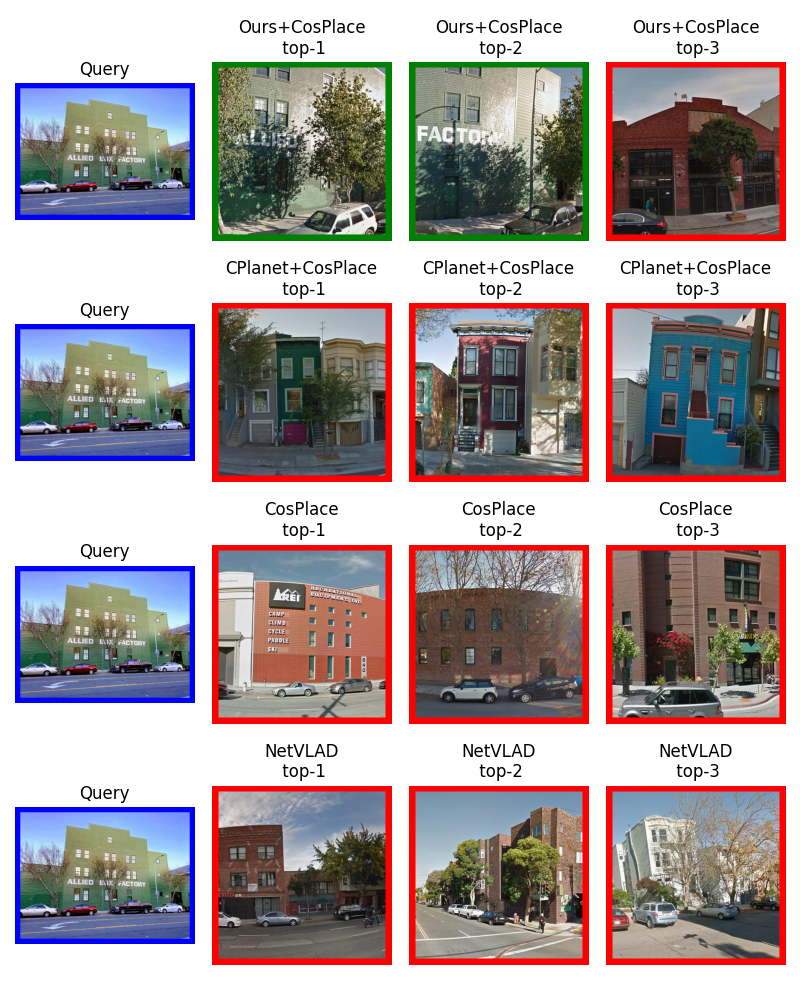}
    \end{subfigure}
    
    \caption{\textbf{Qualitative results} using different pipelines on challenging queries.}
    \label{fig:q_1}
    \vspace{-0.3cm}
\end{figure}

\begin{figure}
    \begin{center}
    \includegraphics[width=\columnwidth]{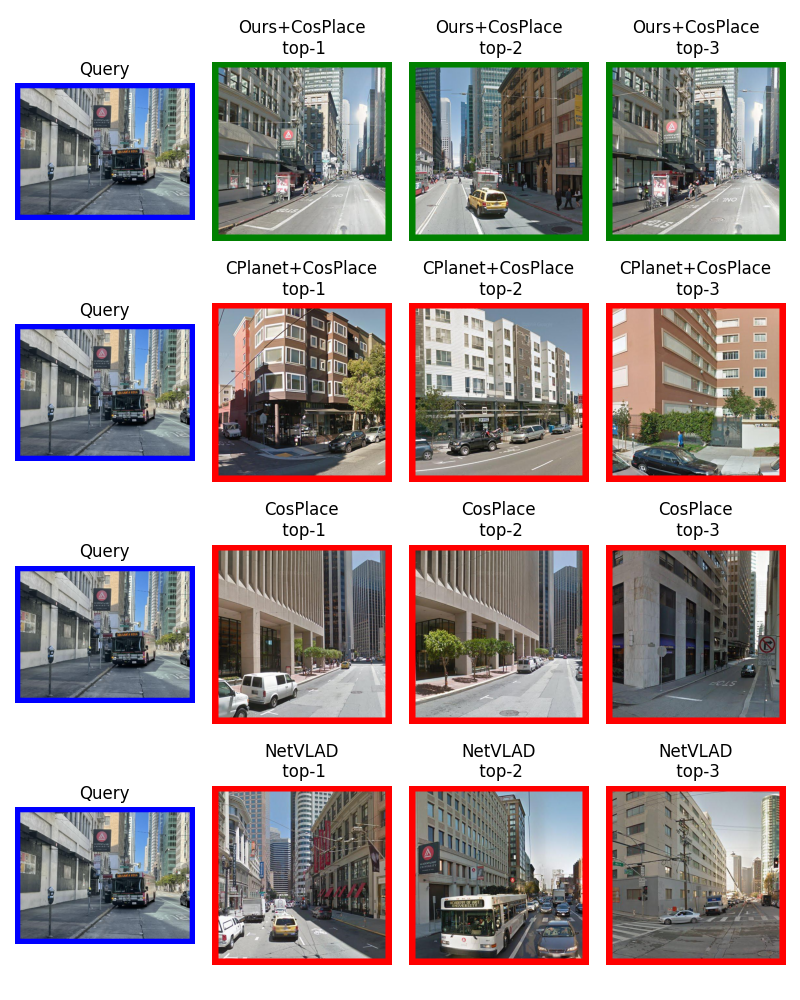}
    \end{center}
    \caption{\textbf{Qualitative results} using different pipelines on challenging queries.}
    \label{fig:q_2}
    \vspace{-0.3cm}
\end{figure}

\myparagraph{Qualitative results}
In \cref{fig:q_1,fig:q_2} we show some qualitative results of challenging queries and the retrieved Top-3 candidates by some retrieval-only methods (namely CosPlace and NetVLAD) and by some classification-retrieval pipelines (using respectively our D\&C and CPlanet as classification modules).

\begin{figure}
    \begin{center}
    \includegraphics[width=0.65\columnwidth]{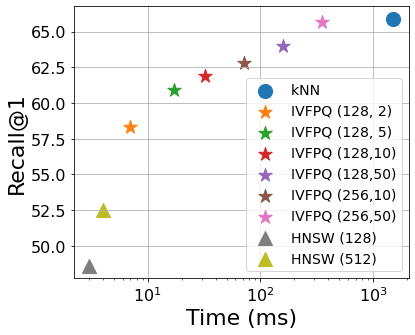}
    \end{center}
    \caption{\textbf{Comparisons of best-performing Approximate Nearest Neighbor search algorithms.} We show only the pareto-optimal results, which are computed with an Inverted File Index with Product Quantization (IVFPQ) \cite{Jegou_2011_productQ} and Hierarchical Navigable Small Worlds (HNSW) \cite{Malkov_2020_hnsw}.
    The parameters in parenthesis for IVFPQ indicate the number of subquantizers and the \textit{nprobe}, \ie the number of Voronoi cells to be searched (out of 1000).
    The parameters in parenthesis for HNSW indicates the number of connections each vertex has within the HNSW graph.}
    \label{fig:ann}
\end{figure}

\myparagraph{Approximate Nearest Neighbor Search}
In \cref{fig:ann} we report the results with the best combinations of methods / hyperparameters for our experiments with Approximate Nearest Neighbor search algorithms.
The plot shows only the best performing configurations.
Among other ANNs that we tried are
standard Product Quantization \cite{Jegou_2011_productQ}, Inverted File Indexes (these two methods can be combined in the IVFPQ), and Inverted File MultiIndex \cite{Babenko_2012_inv_multiindex}. We didn't report these results as they performed poorly \wrt their counterparts in the plot. \\
For Table 1 of the main paper we chose two configuration from this pareto-optimal curve, one being optimized for performances and one for speed. For performances, we picked the configuration that grants at least 10x speed, with the maximum performances, and this turned out to be IVFPQ(128,50). For speed, we selected the methods that provided a speedup of at least 100x. This resulted in choosing IVFPQ(128,2) and HNSW(512).

\subsection{Experiments on small datasets}
\label{sec:small_data}

In the main paper we discussed how classification methods are outperformed by retrieval approaches for small datasets due to the lack of enough positives during training. On the other hand, the inference time gap between both procedures loses relevance when dealing with smaller datasets. In Tab. \ref{tab:limitations} it is presented a quantitative analysis on how the proposed methods behave on datasets that are 1000x smaller than SF-XL, covering geographical areas less than $3km^2$ and having half the density of SF-XL.

\begin{table}
\begin{center}
\begin{adjustbox}{width=0.9\linewidth}
\centering
\begin{tabular}{lcc|cccccc}
\toprule
\multirow{3}{*}{Method} & \multicolumn{2}{c}{C-Pitts30k} & \multicolumn{2}{c}{C-Tokyo 24/7}  \\
& \multicolumn{2}{c}{(30k images)} & \multicolumn{2}{c}{(76k images)}\\
\cline{2-3}\cline{4-5}
& LR@1 & Inf. time& LR@1 & Inf. time\\
\hline
\emph{Classification} & & \\
~~PlaNet \cite{Seo_2018_CPlaNet}                  & 31.5 & 12 ms & 19.5 & 12 ms \\
~~HGE \cite{Muller_2018_hierarchical_geolocation} & 33.6 & 15 ms & 22.0 & 15 ms \\
~~CPlaNet \cite{Seo_2018_CPlaNet}                 & 33.0 & 17 ms & 21.5 & 17 ms \\
~~MvMF \cite{Izbicki_2020_earth_geoloc}           & 31.5 & 12 ms & 19.9 & 12 ms \\
\hline
~~\textbf{D\&C (ours)}                            & 40.5 & 12 ms & 33.7 & 12 ms\\
\hline
\emph{Retrieval} & & (kNN time) & & (kNN time)\\
~~NetVLAD \cite{Arandjelovic_2018_netvlad} & 86.1 & 58 ms & 62.2  & 130 ms\\
~~CRN \cite{Kim_2017_crn}                  & 86.3 & 58 ms & 62.8  & 130 ms \\
~~SARE \cite{Liu_2019_sare}                & 87.2 & 58 ms & 74.8  & 130 ms\\
~~SFRS \cite{Ge_2020_sfrs}                 & 88.7 & 58 ms & 78.5  & 130 ms\\
~~GeM \cite{Radenovic_2019_gem}            & 77.9 & 16 ms & 46.4 & 25 ms \\
~~CosPlace                                 & 88.5 & 16 ms & 82.8 & 25 ms \\
\hline
\emph{Mixed pipeline} & & \\
~~\textbf{\ours (ours)} + CosPlace         &  81.9 & 1 ms & 74.9 & 3.5 ms \\
\bottomrule
\end{tabular}
\end{adjustbox}
\end{center}
\caption{\textbf{Comparison of LR@1} of different methods for \textit{Pitts30k} and \textit{Tokyo24/7} using \textit{EfficientNet-B0} as backbone}
\label{tab:limitations}
\end{table}

\section{Baselines Implementation Details}
\label{sec:further_implementation_details}
Although previous works use different partitioning methods of the dataset in classes, we carefully tuned the partitioning hyperparameters to ensure fair comparisons among different methods.
While some methods split the geographical area according to the density of the training points \cite{Weyand_2016_PlaNet, Muller_2018_hierarchical_geolocation} others fix the dimension of the cells into a predefined value and merge them until the number of geographical regions satisfies the desired condition \cite{Seo_2018_CPlaNet}.

The optimal number of classes generated with each method is shown in \cref{tab:supp_classes_per_dataset}, and in the next paragraphs we detail how we empirically found such values for each partitioning method.

\myparagraph{Partitions of HGE, PlaNet and MVMF}
The three methods of PlaNet \cite{Weyand_2016_PlaNet}, Hierarchical Geolocation Estimation (HGE) \cite{Muller_2018_hierarchical_geolocation} and MVMF \cite{Izbicki_2020_earth_geoloc} all use the same partitions, with the only difference that HGE also uses two coarser splits (\emph{medium} and \emph{coarse}) besides the regular partition (\emph{fine}) used by the other two.
The partitions are built using Google S2 Sphere library, and take as input two parameters, namely $\tau_{min}$ and $\tau_{max}$, which define the minimum and maximum number of images within each cell.
We empirically search for the best values for the parameters on the San Francisco eXtra Large (SF-XL) dataset, and we report the results in \cref{tab:supp_hge_partitions}.
We choose the partitions that lead to the best LR@1 using HGE, and, following their implementation, we use the finer HGE partition also as training set for PlaNet and MVMF.
In practice, this leads to a value of $\tau_{min}=100$ and $\tau_{max}=2500$, as shown in \cref{tab:supp_hge_cplanet_hyperparams} (where we also report the value of $\tau$ for other partitions.
Note that we use proportions between different partitions size according to \cite{Muller_2018_hierarchical_geolocation}.

We tuned cells density on SF-XL since it is the most representative dataset for the studied setting. Remembering that these partitioning schemes are based on keeping cell density constant, to extend the comparison to the other adopted datasets (C-Pitts30k, C-Tokyo24/7), we scaled $\tau_{min}$ and $\tau_{max}$ according to the relative density of the other datasets with respect to SF-XL.
In our method, instead, the partitioning only depends on the desired granularity of localization, so we kept the same $20m$ cells across all datasets.


\begin{table}
\begin{adjustbox}{width=\linewidth}
\centering
\begin{tabular}{lccc}
\toprule
Partition method & SF-XL & C-Pitts30k & C-Tokyo 24/7 \\
\hline
PlaNet / MVMF & 65k & 486 & 1840 \\
HGE & 19k / 35k / 65k & 158 / 272 / 486 & 508 / 961 / 1840 \\
CPlaNet & 54k & 369 & 1236 \\
Ours & 114k & 687 & 2492 \\
\bottomrule
\end{tabular}
\end{adjustbox}
\caption{\textbf{Number of classes in different datasets} using different partitioning methods.
}
\label{tab:supp_classes_per_dataset}
\end{table}

\begin{table}
\centering
\begin{adjustbox}{width=0.6\linewidth}
\begin{tabular}{ccc|c}
\toprule
\multicolumn{3}{c}{HGE Num. Classes}\\

coarse & medium & fine & LR@1\\
\hline
65.3k & 119k  & 200k   & 19.0 \\
35.0k & 65.3k & 119.0k & 21.2\\
18.5k & 35.0k & 65.3k  & 27.0\\
 9.4k & 18.5k & 35.0k  & 25.3\\
 3.8k & 9.4k  & 18.5k  & 19.2\\
 1.8k & 3.8k  & 9.4k   & 10.6\\

\bottomrule
\end{tabular}
\end{adjustbox}
\caption{\textbf{Results with different partitions} using HGE on SF-XL.}
\label{tab:supp_hge_partitions}
\end{table}

\myparagraph{Partitions of CPlaNet}

\begin{table}
\centering
\begin{adjustbox}{width=0.8\linewidth}
\centering
\begin{tabular}{lccc}
\toprule
hyperparams & fine & HGE-medium & HGE-coarse \\
\hline
$\tau_{min}$ & 100 & 100 & 100 \\
$\tau_{max}$ &2500 & 5000 & 10000\\
\bottomrule
\end{tabular}
\end{adjustbox}
\caption{\textbf{Chosen hyperparameters} for previous methods partitioning. Note that Planet, HGE-fine and MvMF use the same partitioning. }
\label{tab:supp_hge_cplanet_hyperparams}
\end{table}

Regarding CPlaNet's \cite{Seo_2018_CPlaNet} partitions, we carefully followed the authors' implementation:
we created five \emph{geoclass sets} for each of the experiments, where \emph{geoclass set\textsubscript{1}} and \emph{geoclass set\textsubscript{2}} evaluate the proximity distance using only the geographical and visual properties of the images respectively, while the remaining \emph{geoclass sets} were generated by considering the distance as a stochastic linear combination of these two modalities.
We refer the reader to their paper for more details about how each \emph{geoclass set} is formed.
In their method, an additional hyperparameter is the number of classes in each \emph{geoclass set} (\ie their partition algorithm stopping condition).
Finally, at inference time, the granularity considered for prediction is given by the intersections of the 5 \emph{geoclass sets}.
In \cref{tab:supp_cplanet_partitions}
we report results using different values for each and using the same parameters $\alpha$ and $\beta$, which define the differences between the 5 \emph{geoclass sets}. The table also reports in the first column the number of distinct cells obtained by the intersection of the different partitions.
Also in this case we choose from the table the split which gave the best results for LR@1.

To export these hyperparameters to the other datasets, we kept the same average size of the cells in each \emph{geoclass set}.

\begin{table}
\centering
\begin{adjustbox}{width=\linewidth}
\begin{tabular}{cccccc|c}
\toprule
& \multicolumn{5}{c}{Cells per geoclass}\\
\# classes & gcs 1 & gcs 2 & gcs 3 & gcs 4 & gcs 5 & LR@1\\
\hline
58233 & 30k & 30k & 39k & 36k & 33k & 27.6 \\ 
54144 & 20k & 20k & 26k & 24k & 22k & 27.7\\
47412 & 10k & 10k & 13k & 12k & 11k & 25.7\\

\bottomrule
\end{tabular}
\end{adjustbox}
\caption{\textbf{CPlaNet preliminary results} on SF-XL. 
}
\label{tab:supp_cplanet_partitions}
\end{table}

\end{document}